%% file: main.tex
\pdfoutput=1

\documentclass[11pt]{article}

\usepackage[final]{acl2023}

\usepackage{times}
\usepackage{latexsym}

\usepackage[T1]{fontenc}

\usepackage[utf8]{inputenc}

\usepackage{microtype}

\usepackage{inconsolata}

\usepackage{booktabs}
\usepackage{linguex}
\usepackage{float}
\usepackage{soul}
\usepackage{colortbl}
\usepackage{multirow}
\usepackage{pgfplots}
\usepackage{color}
\pgfplotsset{compat=newest}
\usetikzlibrary{shapes,arrows}
\usepackage{enumitem}
\setlist{nosep}

\tikzstyle{block} = [rectangle, draw, fill=TolSand!80, text centered, rounded corners, minimum height=2em]
\tikzstyle{tweet} = [rectangle, draw, fill=TolCyan!60,
    text width=30em, text centered, rounded corners, minimum height=0.5em]
\tikzstyle{line} = [draw, -latex']
\tikzstyle{cloud} = [draw, ellipse,fill=TolRose!70, node distance=0.5cm,minimum height=1em]
\graphicspath{{figures/}}

\definecolor{TolIndigo}{HTML}{332288}
\definecolor{TolCyan}{HTML}{88CCEE}
\definecolor{TolTeal}{HTML}{44AA99}
\definecolor{TolGreen}{HTML}{117733}
\definecolor{TolOlive}{HTML}{999933}
\definecolor{TolSand}{HTML}{DDCC77}
\definecolor{TolRose}{HTML}{CC6677}
\definecolor{TolWine}{HTML}{882255}
\definecolor{TolPurple}{HTML}{AA4499}
\definecolor{TolPaleGrey}{HTML}{DDDDDD}

\title{Counterfactual Probing for the Influence of Affect and Specificity\\on Intergroup Bias}

\author{
    Venkata S\ Govindarajan\qquad Kyle Mahowald\qquad David I. Beaver\qquad Junyi Jessy Li\\
    Department of Linguistics\\
    The University of Texas at Austin\\
    \texttt{\{venkatasg, mahowald, dib, jessy\}@utexas.edu}
  }

\begin{document}

\maketitle

\input{sections/00-abstract.tex}

\section{Introduction}
\label{sec:introduction}
\input{sections/01-intro.tex}

\section{Background}
\label{sec:background}
\input{sections/02-background.tex}

\section{Data \& Experiments}
\label{sec:experiments}
\input{sections/02-experiments.tex}

\section{Results \& Analysis}
\label{sec:results}
\input{sections/03-results.tex}

\section{Conclusion}
\label{sec:conclusion}
\input{sections/04-conclusion.tex}

\section*{Limitations}
\label{sec:limitations}
\input{sections/99-limitations.tex}

\section*{Ethics Statement}
\label{sec:Ethics Statement}
\input{sections/99-ethics.tex}

\section*{Acknowledgements}
This research is partially supported by Good Systems\footnote{\url{https://goodsystems.utexas.edu}}, a UT Austin Grand Challenge to develop responsible AI technologies, and NSF grants IIS-2145479, IIS-2107524. We acknowledge the Texas Advanced Computing Center (TACC)\footnote{\url{https://www.tacc.utexas.edu}} at UT Austin and AWS for many of the results within this paper. Kyle Mahowald was funded in part by NSF Grant 2104995.

\bibliography{references}
\bibliographystyle{acl_natbib}

\appendix
\section*{Appendix}
\label{sec:appendix}
\input{sections/99-appendix.tex}

\end{document}

%% file: sections/00-abstract.tex
\begin{abstract}
While existing work on studying bias in NLP focues on negative or pejorative language use, ~\citet{govindarajan-etal-2023-people} offer a revised framing of bias in terms of intergroup social context, and its effects on language behavior. In this paper, we investigate if two pragmatic features (specificity and affect) systematically vary in different intergroup contexts --- thus connecting this new framing of bias to language output. Preliminary analysis finds modest correlations between specificity and affect of tweets with supervised intergroup relationship (IGR) labels. Counterfactual probing further reveals that while neural models finetuned for predicting IGR reliably use affect in classification, the model's usage of specificity is inconclusive.
\end{abstract}

%% file: sections/01-intro.tex
Most work on bias in NLP only considers negative or pejorative language use~\citep{kaneko-bollegala-2019-gender, sheng-etal-2019-woman, webson-etal-2020-undocumented, Pryzant_DiehlMartinez_Dass_Kurohashi_Jurafsky_Yang_2020, sheng-etal-2020-towards}. While recent work has delved into implicit bias~\citep{Rashkin2015ConnotationFA, Sap2017ConnotationFO, sap-etal-2020-social}, they are still limited as they rely on identifying specific demographic dimensions or an individual's intent. Crucially, language production is still taken to be `unbiased' by default. Research in social psychology suggests a different framing of bias that encompasses all language use --- we can analyze bias as changes in (language) behavior reflecting shifting social dynamics~\citep{van2009society}. Under this view, all the language we produce is biased, with the nature of the bias determined by the social relationships between the speaker and target. Inspired by this idea, \citet{govindarajan-etal-2023-people} proposed a new framing of bias by modeling intergroup relationships (IGR, in-group and out-group) in interpersonal English language tweets, potentially capturing more subtle forms of bias. This framing raises a question: \emph{which linguistic features vary systematically in different intergroup contexts?}

The Linguistic Intergroup Bias ~\cite[LIB;][]{maass_language_1989,maass_linguistic_1999} hypothesis offers some clues towards linguistic features that change with shifting intergroup contexts. LIB speculates that socially desirable in-group behaviors and socially undesirable out-group behaviors are encoded at a higher level of abstraction. The theory however relies on a restricted definition of abstractness that relies solely on predicates, and an ad-hoc analysis of `social desirability' that doesn't permit large-scale analysis. We can do better by using two well-defined pragmatic features: \textbf{specificity}~\citep{li_discourse_2017} is a pragmatic feature of text that measures the level of detail (similar to abstract--concrete axis), while \textbf{affect} is a feature that measures the attitude of a speaker towards their target~\citep{sheng-etal-2019-woman} in an utterance (analogous to social desirability).

Specificity and affect are analogous to the LIB axes of language variation that are easy to annotate and compute. Furthermore, specificity is a more \emph{general} property than abstractness in the LIB --- specificity is a property of the whole sentence rather than just the predicate. Thus, our study focuses on \textbf{intergroup bias} more generally, rather than the narrow parameterization of the LIB. Similar to the LIB, our formulation of intergroup bias predicts that positive affect in-group utterances and negative affect out-group utterances are encoded with \emph{lower specificity} (i.e. more generally). Tables~\ref{tab:lib} and \ref{tab:ib} compare the predicted language variation between the LIB and our formulation. 

In this work, we perform the first large-scale study of linguistic differences in intergroup bias by analyzing its nature in the corpus of English tweets from \citet{govindarajan-etal-2023-people}, which makes use of naturally occurring labels for in-group vs. out-group. This distinguishes us from existing work in LIB which mostly relies on artificial responses from participants in studies, rather than natural language use in the wild. To bolster our probing investigation, we also explore it causally: exploiting the quantitative nature of our formulation to study if a neural model finetuned for IGR prediction uses pragmatic features such as specificity and affect in its decision-making process through counterfactual probing techniques~\citep{ravfogel-etal-2021-counterfactual}.

To summarize our findings, we find a modest positive correlation between affect and IGR in our data, with a positive causation effect as well --- making a tweet's affect more positive makes it more likely to be in-group regardless of its specificity. We find no correlation between specificity and IGR in our data. Surprisingly, we discover a causal effect of low specificity on IGR prediction that is uniform across affect, but none for high specificity. We hypothesize that this could be because of damage to the underlying language model, but we leave further investigation to future work. We release our code and data at \href{https://github.com/venkatasg/intergroup-probing}{\texttt{github.com/venkatasg/intergroup-probing}}.

%% file: sections/02-background.tex
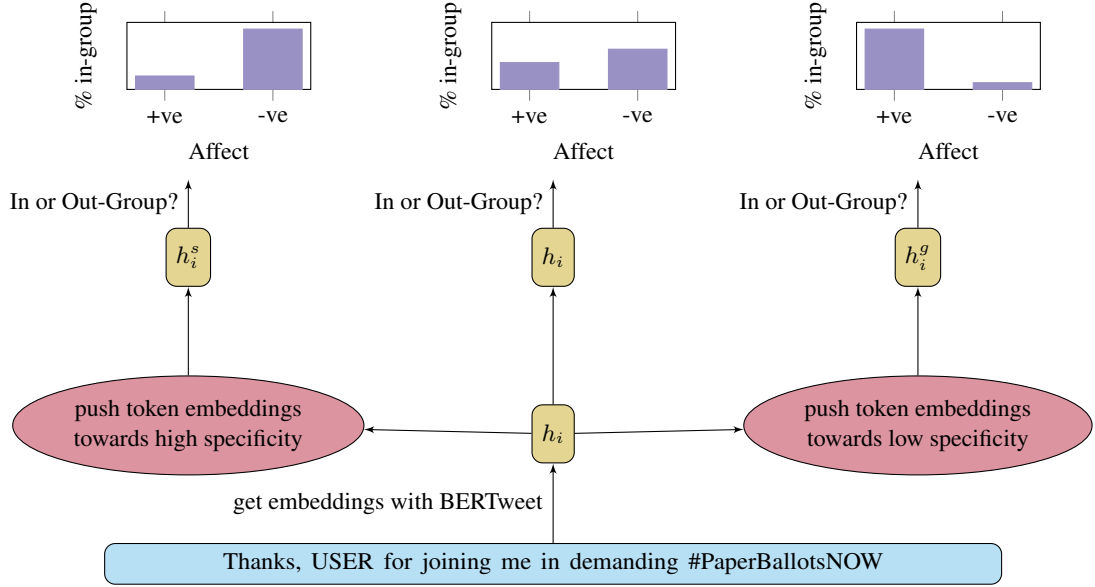
\begin{figure*}[t]
    \centering
    \input{figures/flowchart.tex}
    \caption{Flowchart describing the specificity intervention experiment and expected results.}
    \vspace{-5mm}
    \label{fig:flowchart}
\end{figure*}

\input{tables/lib.tex}

\paragraph{Intergroup bias} The  Linguistic Intergroup Bias  (LIB) theory~\cite{maass_language_1989,maass_linguistic_1999} tries to explain how stereotypes are transmitted and persist in communication by hypothesizing that socially desirable in-group behaviors and socially undesirable out-group behaviors are encoded at a higher level of abstraction . The LIB has been reproduced in various psychological experiments and analyses~\cite{Anolli2006LinguisticIB, gorham_news_2006}; it has also been used as an indicator for a speaker’s prejudicial attitudes~\cite{hippel_linguistic_1997}, and racism~\cite{schnake_modern_1998}. 

Table~\ref{tab:lib} describes the LIB asymmetry and the parameters used. As stated earlier, the LIB relies on ad-hoc and hand-coded concepts such as `social desirability' and abstractness of predicates~\citep{semin_cognitive_1988} . Our proposed experiments \emph{generalize} beyond the LIB by utilizing parameters that are easily computable, and are a function of the whole utterance. We also build upon the dataset and work in \citet{govindarajan-etal-2023-people}, which is the first large-scale analysis of intergroup bias on naturally occurring speech.

\paragraph{Specificity} Specificity is a pragmatic concept of text that measures the level of detail and involvement of concepts, objects and events. \citet{louis-nenkova-2011-text} introduced the first dataset and model for sentence specificity prediction, and in later work \citet{li_discourse_2017} illustrated the role of specificity in discourse coherence. Furthermore, \citet{gao_predicting_2019} expanded the scope of specificity analysis from the news domain to social media.

\input{tables/ib.tex}

\paragraph{Affect} There is a wealth of work studying emotions and sentiment in social media text~\citep{mohammad-2012-emotional,6406313, Mohammad2015UsingHT, abdul-mageed-ungar-2017-emonet,desai-etal-2020-detecting,demszky-etal-2020-goemotions}. \citet{govindarajan-etal-2023-people} introduced the first dataset annotated for \emph{interpersonal} emotion (defined as only emotions expressed towards or in connection with a target), using the Plutchik wheel~\citep{plutchik1980general, plutchik2001nature} as a framework. While fine-grained, this approach isn't amenable to the experimentation we propose easily. Inspired by the concept of \emph{regard} by a speaker towards a demographic in an utterance~\citep{sheng-etal-2019-woman}, we introduce annotations for a coarse-grained feature we term \emph{affect} that estimates how a speaker feels towards the target they mentioned in an interpersonal utterance.

Table~\ref{tab:ib} describes the intergroup language variation as hypothesized in our experimentation, using specificity and affect. Analogous to LIB, our hypothesis is that positive affect utterances directed at in-group individuals, and negative affect utterances directed at out-group individuals are encoded with \emph{lower specificity}.

\paragraph{AlterRep} AlterRep~\citep{ravfogel-etal-2021-counterfactual} is a probing technique that tests if a neural network \emph{uses} a property, rather than just testing if the model's learned representations correlate with the property. The method uses Iterative Nullspace Projection~\citep[INLP;][]{ravfogel-etal-2020-null} to iteratively train a linear classifier on the model's internal representations to pick out a particular feature, using the parameters learned by the classifier to intervene on the embedding and alter it in a systematic way. The AlterRep method based on INLP has been used to probe for syntactic phenomena such as subject-verb number agreement~\citep{ravfogel-etal-2021-counterfactual}. To our knowledge, ours is the first work probing if a model learns and uses higher-level pragmatic features like affect and specificity using AlterRep.

%% file: figures/flowchart.tex
\begin{tikzpicture}[node distance = 0.1\textwidth, auto]
    \node (middle-plot) {
        \begin{tikzpicture}
            \begin{axis}[ybar, ymin=0,ymax=100,
                         symbolic x coords={+ve,-ve},xtick=data,  /pgf/bar width=2em, width=0.25\textwidth,height=0.1\textheight, tick label style={font=\small}, yticklabels=None, enlarge x limits=0.35, ytick style={draw=none}, xlabel={\small Affect}, ylabel={\small\% in-group}]
                \addplot+[ybar, color=TolIndigo!50]
                    coordinates {(+ve,40) (-ve,60)};
            \end{axis}
        \end{tikzpicture}
    };
    \node [left of=middle-plot, xshift=-0.2\textwidth] (left-plot) {
        \begin{tikzpicture}
            \begin{axis}[ybar, ymin=0,ymax=100, ylabel={Hate Speech \%},
                         symbolic x coords={+ve,-ve},xtick=data,  /pgf/bar width=2em, width=0.25\textwidth,height=0.1\textheight, tick label style={font=\small},yticklabels=None, enlarge x limits=0.35, ytick style={draw=none}, xlabel={\small Affect}, ylabel={\small\% in-group}]
                \addplot+[ybar, color=TolIndigo!50]
                    coordinates {(+ve,20) (-ve,90)};
            \end{axis}
        \end{tikzpicture}
    };
    \node [right of=middle-plot, xshift=0.2\textwidth] (right-plot) {
        \begin{tikzpicture}
            \begin{axis}[ybar, ymin=0,ymax=100,
                         symbolic x coords={+ve,-ve},xtick=data, /pgf/bar width=2em, width=0.25\textwidth,height=0.1\textheight, tick label style={font=\small},yticklabels=None, enlarge x limits=0.35, ytick style={draw=none}, xlabel={\small Affect}, ylabel={\small\% in-group}]
                \addplot+[ybar, color=TolIndigo!50]
                    coordinates {(+ve,90) (-ve,10)};
            \end{axis}
        \end{tikzpicture}
    };
    \node [block, below of=middle-plot, yshift=-0.03\textheight] (bert-1) {\small $h_i$};
    \node [block, left of=bert-1, xshift=-0.2\textwidth] (bert-pos) {\small $h_i^s$};
    \node [block, right of=bert-1, xshift=0.2\textwidth] (bert-neg) {\small $h_i^g$};
    
    \node [block, below of=bert-1, yshift=-0.03\textheight] (bert) {\small  $h_i$};
    \node [cloud, below of=bert-pos, yshift=-0.07\textheight] (pos) {\shortstack{\small  push token embeddings\\\small  towards high specificity}};
    \node [cloud, below of=bert-neg, yshift=-0.07\textheight] (neg) {\shortstack{\small  push token embeddings\\\small  towards low specificity}};
    
    \node [tweet, below of=bert, yshift=-0.005\textheight] (tweet) {\small Thanks, \@USER for joining me in demanding \#PaperBallotsNOW};
    
    \path [line] (tweet) -- node {\small  get embeddings with BERTweet}(bert);
    \path [line] (bert) -- (bert-1);
    \path [line] (pos) -- (bert-pos);
    \path [line] (neg) -- (bert-neg);
    \path [line] (bert) -- (pos);
    \path [line] (bert) -- (neg);
    \path [line] (bert-1) -- node[pos=0.5,left] {\small In or Out-Group?}(middle-plot);
    \path [line] (bert-neg) -- node[pos=0.5,left] {\small In or Out-Group?}(right-plot);
    \path [line] (bert-pos) -- node[pos=0.5,left] {\small In or Out-Group?}(left-plot);
\end{tikzpicture}

%% file: tables/lib.tex
\begin{table}[t]
    \centering
    \small
    \begin{tabular}{lll}
     \toprule
        \textbf{} & \textbf{In-group} & \textbf{Out-group} \\ \midrule
        socially desirable & abstract & concrete \\ \midrule
        socially undesirable & concrete & abstract \\ 
    \bottomrule
    \end{tabular}
    \caption{Predicted language variation in the LIB.}
    \label{tab:lib}
    \vspace{-5mm}
\end{table}

%% file: tables/ib.tex
\begin{table}[t]
    \centering
    \small
    \begin{tabular}{lll}
     \toprule
        \textbf{} & \textbf{In-group} & \textbf{Out-group} \\ \midrule
        positive affect & low specificity &  high specificity \\ \midrule
        negative affect & high specificity & low specificity \\ 
    \bottomrule
    \end{tabular}
    \caption{Predicted language variation in our more general formulation, using specificity and affect}
    \label{tab:ib}
    \vspace{-5mm}
\end{table}

%% file: sections/02-experiments.tex
\subsection{Data \& Annotations}
\label{subsec:data}

We use the same dataset of tweets from \citet{govindarajan-etal-2023-people}, which consists of tweets by members of US Congress that @-mention other members in the same tweet, with `found-supervision' for the IGR labels of every tweet. A tweet is in-group if it is targeted at another member of the same party as the writer of the tweet, else it is out-group.

\paragraph{Affect}
We build upon the dataset's fine-grained annotations for interpersonal emotion by adding annotations for affect. We presented annotators on Mechanical Turk with tweets from our dataset with the target mention masked (with the placeholder \@Doe, to minimize potential biases of the annotator), and asked the following questions:

\begin{enumerate}
	\item[a.] How does the writer feel in general about \@Doe? \emph{warmly, coldly, neutral, mixed}
	\item[b.] How does the writer feel in general about \@Doe's actions/behavior? \emph{approval, disapproval, neutral, mixed}
\end{enumerate}

Annotators are given the option to select one of the 4 options listed above for each question. For each tweet, we collect annotations from 3 annotators, obtaining an aggregate label for each question by majority vote. We report an inter-annotator agreement score ~\citep[Fleiss's kappa;][]{fleiss1971measuring} of 0.53 for the first question, and 0.56 for the second.

We derive a binary affect label ($\pm 1$) from our annotations using a simple rule: If the writer of a tweet is deemed to either feel warmly towards the target, or if they approve of the target's actions, the affect is set to be positive; else it is set to be negative. An analysis of our collected annotations on the data shows that there is a small positive (Pearson's) correlation ($r$=$0.2$, $p<0.001$) between binary affect and IGR.

\paragraph{Specificity} Specificity of the tweets in the dataset are calculated using the specificity  prediction tool from \citet{gao_predicting_2019}. Their specificity predictor is trained on tweets, and uses surface lexical features, as well as syntactic, semantic and distributional features to calculate a specificity score between 1 and 5. We note that on our dataset, there was \emph{no correlation between specificity and IGR} ($r$=$-0.07$, $p<0.001$), unlike affect. On further inspection of our dataset, we find that tweets with very high/low specificity scores \citep[gathered by excluding specificity scores between 3 and 4, similar to excluding the middle in][]{gelman2009} have a small but statistically significant negative correlation with IGR labels ($r$=$-0.13$, $p<0.001$).

\subsection{Interventions}
\label{subsec:interventions}

\paragraph{Model} We use BERTweet~\citep{nguyen_bertweet_2020}, a language model pre-trained on 850M English tweets, the same model used in \citet{govindarajan-etal-2023-people}. All intervention experiments are carried out with the best performing \emph{finetuned} version of this model --- where the model is finetuned on the task of predicting IGR labels. The input to the model is only the tweet with no other context, and the target masked with a placeholder \texttt{@USER}.

We use the model's representations from layer 11 for the INLP procedure since it shows the most reliable effects. INLP \citep{ravfogel-etal-2020-null} works by learning a series of linear classifiers on the representations from an encoder. In each iteration, the embeddings are projected onto the intersection of nullspaces of the classifiers learned so far, meaning the information used by the existing classifiers is removed from the model. Every subsequent classifier we learn removes more information of the property of interest from the model's representations.  We find that higher layers offer a good balance between feature extractability and language model stability (see Appendix~\ref{app:lm-acc}) for our features. 

After training INLP, AlterRep uses the classifier's decision space to project model embeddings into a null component that contains no information from the feature of interest, and an orthogonal component, that contains all the information from the feature of interest. These two components thus enable us to perform the counterfactual intervention --- pushing model embeddings towards having more, or less, of a particular property. When AlterRep uses INLP classifiers with more iterations, the strength of the intervention is greater. Figure~\ref{fig:flowchart} offers an illustration of our intervention experiment on specificity, and the expected results.

\paragraph{Affect} Using the binary affect labels we derived from annotations that we described in \textsection~\ref{subsec:data}, we perform interventions to test if the model uses affect causally in its decision. We sample 3 tokens at random from each sentence in the training and validation split of our dataset, train an iterative linear classifier on the model's representations of these tokens using INLP (against the affect label of the tweet), and use the decision boundary learned by the classifier to intervene by pushing model representations to have more positive affect or have more negative affect. We set the hyperparameter $\alpha$ in AlterRep to 4.

\paragraph{Specificity} The INLP classifier for specificity is learned using the same procedure as for affect. We train the classifier on only the tweets with high and low specificity scores in our dataset (scores below 3 and above 4; scores taken from the specificity prediction tool in \citet{gao_predicting_2019}), excluding the middle to ensure effective learning of the decision boundary~\citep{gelman2009}. Thus, we are effectively pushing the model representations to have high or low specificity. For both affect and specificity, once the INLP classifier is learned, we perform the intervention on a random subset of 30\% of the tokens of a tweet (to control for tweet length). We also report the results of random interventions as a control, where random interventions are generated by sampling from a standard gaussian instead of using the decision matrix generated by INLP.

\paragraph{Hypotheses} We report the percentage of tweets in the test split of our dataset that are predicted to be in-group by our classifier model with increasing strength of the intervention (number of INLP iterations, 0 being pre-intervention). Thus, we have the following hypotheses on the effects of our intervention on the data based on our intergroup bias framework described in Table~\ref{tab:ib}:

\begin{enumerate}
	\item Interventions towards positive affect should induce the model to predict low specificity  tweets to be in-group and high specificity tweets to be out-group, while interventions towards negative affect should affect the model conversely.
	\item Interventions towards higher specificity should induce the model to predict positive affect tweets as out-group and negative affect tweets as in-group, while interventions towards lower specificity should affect the model conversely.
\end{enumerate}

%% file: sections/03-results.tex
\begin{figure}[t]
    \centering
    \small
    \input{figures/affect.tex}
    \vspace{-6mm}
    \caption{\% of test set classified as in-group plotted against number of INLP interventions for affect.}
    \vspace{-4mm}
    \label{fig:affect}
\end{figure}
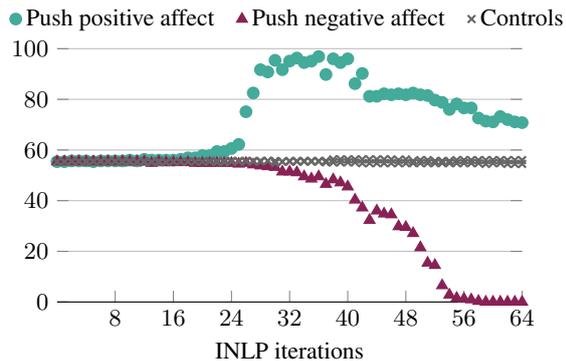
\begin{figure}[t]
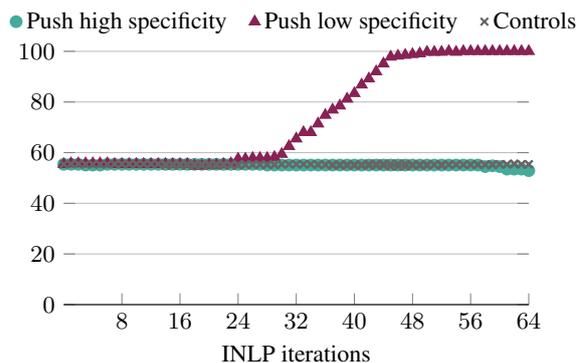

    \centering
    \small
    \include{figures/specificity.tex}
    \vspace{-1cm}
    \caption{\% of test set classified as in-group plotted against number of INLP interventions for specificity.}
    \vspace{-5mm}
    \label{fig:spec}
\end{figure}

The results for the interventions on affect are presented in Figure~\ref{fig:affect}, while those for specificity are presented in Figure~\ref{fig:spec}. Overall, we observe that in both cases, interventions had the same effect on tweets that were annotated with positive affect as they did on tweets with negative affect (and similarly for tweets with high and low specificity) --- so we only show the percentage of \emph{all} tweets in the test split classified as in-group.

\paragraph{Affect} As Figure~\ref{fig:affect} shows, pushing model representations towards having more positive affect causes almost all tweets in the test split of our data to be classified as in-group after 32 iterations of INLP. The randomness after 40 iterations of INLP could be attributed to the underlying RoBERTa language model being destroyed, as the LM Top-100 accuracy plot in Appendix~\ref{app:lm-acc} shows. Pushing the model's representations towards negative affect shows the inverse effect as expected, although the nature of the drop appears different. We hypothesize that this is because most of the tweets in our dataset (75.2\%) have positive affect. An intervention pushing the representations towards negative affect would be slower and require stronger intervention forces, which is borne out in Figure~\ref{fig:affect}.

\paragraph{Specificity} Figure~\ref{fig:spec} shows that pushing model representations towards being more specific has no effect on model behavior and is indistinguishable from the control; but pushing towards lower specificity has a noticeable effect --- interventions after 48 iterations of iNLP lead to all the data being predicted as in-group. Our hypothesis states that general language is more likely in positive affect in-group contexts; however we find no difference in the model's behavior on positive versus negative affect tweets as reported earlier.

Overall our findings indicate that while the model does use affect towards making its decision on the interpersonal group relationship prediction task (albeit uniformly across specificity), it doesn't use specificity as we had predicted. The discrepancy between high and low specificity interventions could be because the average specificity of tweets in our training data is 3.49 ($\sigma=0.54$) --- meaning that interventions towards lower specificity act in opposition to most of our data in representation space. But these results requires further investigation to understand them better.

\paragraph{Qualitative error analysis} Digging into the results further, we wanted to investigate if the interventions function the way we wanted them to. We analyzed the tokens that the model predicts before and after intervention for example \ref{ex:qual}. Firstly, finetuning the model for IGR prediction leads to degradation in LM abilities --- a vanilla model predicts \emph{birthday, anniversary} for the masked token in \ref{ex:qual}, but the finetuned model predicts nonsensical tokens like \emph{sworn, opport\_\_} even before any interventions. Pushing towards negative affect causes it to predict tokens with negative connotations (\emph{killing, ass, opposition}), but degrades the underlying LM even further. The specificity interventions are especially hard to interpret due to the semantically and syntactically implausible tokens being selected (\emph{opport\_\_, mug\_\_, ask\_\_}) 

\ex.\label{ex:qual} Happy <mask> @USER! I got you a new bill: \#IIOA

While some of the interventions push the model's predictions to be in the general lexical space desired (which probably explain the affect intervention results), the lack of contextual fit due to LM degradation may explain the inconclusive results, and lack of interaction between affect and specificity.

%% file: figures/affect.tex
\begin{tikzpicture}
	\begin{axis}[
		xlabel={INLP iterations},
        axis x line*=bottom,
        y axis line style= { draw opacity=0 },
        ymajorgrids=true,
        xtick={8,16,24,32,40,48,56,64},
        enlarge x limits=-1,
        ymin=0,
        ymax=100,
        height=0.2\textheight,
        width=\columnwidth,
        legend columns=-1,
        legend style={font=\small, at={(0.5,1.2)},anchor=north, draw=none, /tikz/every even column/.append style={column sep=0.03\columnwidth}},
          legend entries={Push positive affect,
                          Push negative affect,
                          Controls
                          }
    ]
    \addplot[only marks, thick, color=TolTeal, mark=*, mark size=2pt] table [x=inlp, y=posaff, col sep=tab] {data/test.tsv};
    
    \addplot[only marks, thick, color=TolWine, mark=triangle*, mark size=2pt] table [x=inlp, y=negaff, col sep=tab] {data/test.tsv};
    
    \addplot[only marks, thick, color=black!60, mark=x, mark size=2pt] table [x=inlp, y=posaff-control, col sep=tab] {data/test.tsv};
    
    \addplot[only marks, thick, color=black!60, mark=x, mark size=2pt] table [x=inlp, y=negaff-control, col sep=tab] {data/test.tsv};
	\end{axis}
\end{tikzpicture}

%% file: figures/specificity.tex
\begin{tikzpicture}
	\begin{axis}[
		xlabel={INLP iterations},
        axis x line*=bottom,
        y axis line style= { draw opacity=0 },
        ymajorgrids=true,
        xtick={8,16,24,32,40,48,56,64},
        enlarge x limits=-1,
        ymin=0,
        ymax=100,
        height=0.2\textheight,
        width=\columnwidth,
        legend columns=-1,
        legend style={font=\small, at={(0.5,1.2)},anchor=north, draw=none, /tikz/every even column/.append style={column sep=0.03\columnwidth}},
          legend entries={Push high specificity,
                          Push low specificity,
                          Controls
                          }
    ]
    \addplot[only marks, thick, color=TolTeal, mark=*, mark size=2pt] table [x=inlp, y=posspec, col sep=tab] {data/test.tsv};
    
    \addplot[only marks, thick, color=TolWine, mark=triangle*, mark size=2pt] table [x=inlp, y=negspec, col sep=tab] {data/test.tsv};
    
    \addplot[only marks, thick, color=black!60, mark=x, mark size=2pt] table [x=inlp, y=posspec-control, col sep=tab] {data/test.tsv};
    
    \addplot[only marks, thick, color=black!60, mark=x, mark size=2pt] table [x=inlp, y=negspec-control, col sep=tab] {data/test.tsv};
	\end{axis}
\end{tikzpicture}

%% file: sections/04-conclusion.tex
Studying bias in language use through an interpersonal lens opens up new questions, such as which linguistic features vary systematically in changing interpersonal contexts. We perform a correlational and causal analysis of two pragmatic features, specificity and affect, on a dataset of interpersonal tweets in English, to establish how they influence intergroup relationship prediction. We find modest correlations between our features and IGR labels, while counterfactual probing reveals mixed results. Affect influences IGR prediction causally but without interacting with specificity, while specificity only influences IGR prediction in one direction. 

%% file: sections/99-limitations.tex
Future work must look into the generalizability of the results presented here in other domains of language use, and other languages. While we present the utterances as constituting natural speech by one speaker (the congressperson who sent the tweet), it is likely most congresspeople employ social media teams that help in crafting the language of some of their tweets. However, we believe for the sake of interpersonal group membership, the relationship between the speaker(or speakers) and their target(s) would not be affected.

Techniques like INLP extract information that is linearly extractable. While we've shown that it is possible to extract and manipulate language information using such simple linear techniques, more complex methods like those proposed by \citet{https://doi.org/10.48550/arxiv.2201.12191} might be able to manipulate more non-linearly encoded properties.

The AlterRep procedure, as can be seen in our results and in \citet{ravfogel-etal-2021-counterfactual}, is sensitive to parameters like $\alpha$ and the number of INLP iterations. Picking these parameters is tricky and we have done it in a manner that preserves information in the language model.
It is possible that a different set of settings not explored here could lead to different results.

%% file: sections/99-ethics.tex
For the corpus of tweets on which we performed annotations, we downloaded the tweets using the official Twitter API. In accordance with the Twitter Terms of Service, we release tweet IDs and usernames, but not the tweet text itself. Our dataset was built through crowdsourced annotations on Amazon Mechanical Turk. To ensure annotators were paid a fair wage of at least \$10 an hour, we paid annotators \$0.50 per HIT. Each HIT involved annotating 3 tweets, which we estimate to take on average 3 minutes to complete.

%% file: sections/99-appendix.tex
\section{Implementation}
\label{app:implementation}

We use \texttt{bertweet-base} models from VinAI's Huggingface models  repository, and the \texttt{transformers} package for all of our probing experiments~\citep{wolf-etal-2020-transformers}. Language ID classifiers were trained using \texttt{LinearSVC} classifier from \texttt{sklearn}. For training these classifiers, equal number of tokens from both labels were sampled. We used a batch size of 32, and a maximum sequence length of 128 when performing the intervention experiments. The interpersonal group relationship prediction model was reproduced from \citet{govindarajan-etal-2023-people} using the same experimental settings and hyperparameters for the probing experiments.

\input{tables/datasplit.tex}

\section{Data \& Annotation}
\label{app:annotation}

To obtain reliable annotations, we prequalify annotators using a qualifying task. Annotators were recruited on Mechanical Turk using a qualifying task where they were asked to annotate 6 tweets using the schema detailed in\textsection~\ref{subsec:data}. We restricted the qualification task to annotators living in the USA who had attempted at least 500 HITS and had a HIT approval rate $\geq$ 98\%. After manual inspection, 6 anonymous annotators were qualified for bulk annotation. Each tweet was annotated by three different annotators. To ensure annotators were paid a fair wage of at least 10\$ an hour, we paid annotators \$0.50 per HIT. Each HIT involved annotating 3 tweets, which we estimate to take on average 3 minutes to complete.  In total, 3,033 tweets between 2010 and 2021 were annotated with perceived affect.

\section{Dataset Statistics}
\label{app:data}

We present preliminary statistics for the annotations on the dataset of tweets in Table~\ref{tab:traindevtestsplit}.

\begin{figure}[t]
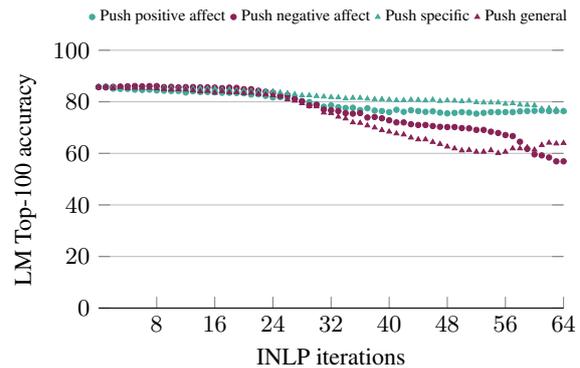

	\centering
    \small
	\include{figures/lmacc.tex}
	\caption{LM Accuracy on train plus validation split for different interventions}
	\label{fig:lmacc}
\end{figure}

\section{LM Accuracy over INLP iterations}
\label{app:lm-acc}

AlterRep directly alters the LM's representations, which inevitably harms the model's internal structure. Figure~\ref{fig:lmacc} shows the LM's top-100 accuracy at predicting randomly masked tokens on our dataset, proving that the interventions are meaningful while still maintaining the LM's integrity.

%% file: tables/datasplit.tex
\begin{table}[t]
	\centering
	\small
	\begin{tabular}{llll}
		\toprule
		\textbf{Affect} & \textbf{Train} & \textbf{Dev} & \textbf{Test} \\ \midrule
		Positive & 1813 & 226 & 242\\
        Negative & 589 & 80 & 83\\
        \bottomrule
	\end{tabular}
	\caption{Distribution of affect in train-dev-test split}
	\label{tab:traindevtestsplit}
\end{table}

%% file: figures/lmacc.tex
\begin{tikzpicture}
	\begin{axis}[
		xlabel={INLP iterations},
		ylabel={LM Top-100 accuracy},
        axis x line*=bottom,
        y axis line style= { draw opacity=0 },
        ymajorgrids=true,
        xtick={8,16,24,32,40,48,56,64},
        enlarge x limits=-1,
        ymin=0,
        ymax=100,
        legend columns=-1,
        height=5cm,
        width=\columnwidth,
        legend style={font=\tiny, at={(0.5,1.2)},anchor=north, draw=none},
          legend entries={Push positive affect,
                          Push negative affect,
                          Push specific,
                          Push general
                          }
    ]
    \addplot[only marks, color=TolTeal, mark=*, mark size=1pt] table [x=inlp, y=posaff, col sep=tab] {data/lmacc.tsv};

    \addplot[only marks, color=TolWine, mark=*,, mark size=1pt] table [x=inlp, y=negaff, col sep=tab] {data/lmacc.tsv};

    \addplot[only marks, color=TolTeal, mark=triangle*, mark size=1pt] table [x=inlp, y=spec, col sep=tab] {data/lmacc.tsv};
    
    \addplot[only marks, color=TolWine, mark=triangle*, mark size=1pt] table [x=inlp, y=general, col sep=tab] {data/lmacc.tsv};
	\end{axis}
\end{tikzpicture}